\RequirePackage{pdf14}
\documentclass{article}

\usepackage{graphicx}
\usepackage{amsmath}
\usepackage{amssymb}
\usepackage{float}
\usepackage{url}
\usepackage{xcolor}
\usepackage{makecell}
\usepackage{cite}
\usepackage{hyperref}
\graphicspath{ {images/} }

\newcommand{\eg}{\emph{e.g.}, }
\newcommand{\ie}{\emph{i.e.}, } 

\usepackage{subcaption} % Table captions
\newcommand{\etal}{\textit{et al.}}

\begin{document}

\title{\LARGE \bf Fast Direct Stereo Visual SLAM}

\author{Jiawei Mo$^{1}$, Md Jahidul Islam$^{2}$, and Junaed Sattar$^{3}$%
\thanks{The authors are with the Department of Computer Science and Engineering, Minnesota Robotics Institute (MnRI), University of Minnesota Twin Cities, Minneapolis, MN, USA.
{Email: \tt\small \{$^{1}$moxxx066, $^{2}$islam034, $^{3}$junaed\}@umn.edu.}}
}
\date{}

\maketitle

\begin{abstract}
We propose a novel approach for fast and accurate stereo visual Simultaneous Localization and Mapping (SLAM) independent of feature detection and matching. We extend monocular Direct Sparse Odometry (DSO) to a stereo system by optimizing the scale of the 3D points to minimize photometric error for the stereo configuration, which yields a computationally efficient and robust method compared to conventional stereo matching. We further extend it to a full SLAM system with loop closure to reduce accumulated errors. With the assumption of forward camera motion, we imitate a LiDAR scan using the 3D points obtained from the visual odometry and adapt a LiDAR descriptor for place recognition to facilitate more efficient detection of loop closures. Afterward, we estimate the relative pose using direct alignment by minimizing the photometric error for potential loop closures. Optionally, further improvement over direct alignment is achieved by using the Iterative Closest Point (ICP) algorithm. Lastly, we optimize a pose graph to improve SLAM accuracy globally. By avoiding feature detection or matching in our SLAM system, we ensure high computational efficiency and robustness. Thorough experimental validations on public datasets demonstrate its effectiveness compared to the state-of-the-art approaches.
\end{abstract}
\section{Introduction}
\label{sec:introduction}

Simultaneous Localization and Mapping (SLAM) has been an active research problem in robotics and computer vision over the past few decades~\cite{cadena2016past,thrun2007simultaneous}. It deals with estimating a robot's instantaneous location by using onboard sensory measurements, \eg LiDAR (Light Detection and Ranging) sensors, cameras, and inertial measurement units (IMU). SLAM is particularly useful where GPS reception is weak such as indoor, urban, and underwater environments. Hence, it has been an essential component in AR/VR~\cite{grasa2013visual}, autonomous driving~\cite{bresson2017simultaneous}, and GPS-denied robotics applications~\cite{weiss2011monocular}. Among existing systems, visual SLAM \cite{fuentes2015visual} is of significant interest because cameras are low-cost passive sensors and thus consume less energy compared to active ones such as sonar or LiDAR. Autonomous mobile robots operating outdoors benefit greatly from the low power consumption of cameras in long-term deployments. 

Visual SLAM systems can be categorized into \emph{feature-based} methods and \emph{direct} methods. The feature-based methods~\cite{klein2007parallel, mur2015orb} detect and match features across frames, and then estimate relative camera motion by minimizing the reprojection error; whereas direct methods~\cite{engel2014lsd, engel2017direct} estimate camera motion by minimizing photometric error directly without feature correspondences. The direct methods demonstrate higher accuracy and robustness over feature-based methods, especially in poorly-textured (less-textured or repetitively-textured) environments \cite{forster2016svo}. As the feature detection and matching algorithms are computationally expensive, sparse direct methods also have the potential to run much faster (\eg $\geq300$ FPS for SVO \cite{forster2016svo}). On the other hand, visual SLAM systems can also be categorized into monocular systems and multi-camera systems. Monocular systems~\cite{klein2007parallel,engel2014lsd,mur2015orb,engel2017direct} cannot estimate the metric \textit{scale} of the environment which multi-camera systems are able to. Multi-camera systems usually achieve higher accuracy and robustness; among these, stereo systems~\cite{mur2017orb,engel2015large,wang2017stereo} are particularly popular for their simplicity and accessibility.

\begin{figure}[t]
    \centering
    \includegraphics[width=\textwidth]{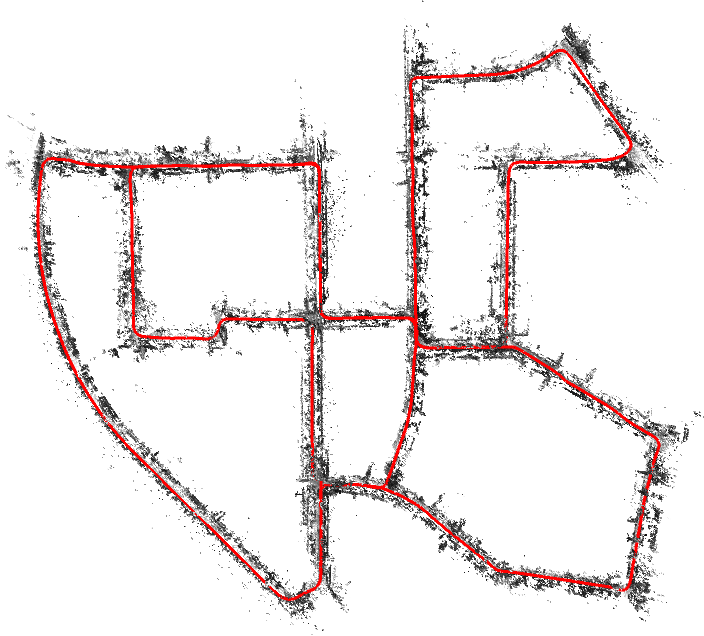}
    \caption{The estimated trajectory and reconstructed environment by the proposed method on KITTI sequence $00$.}
    \label{fig:kitti00}
\end{figure}

Most existing stereo visual systems use the standard \textit{stereo matching} algorithm~\cite{hartley2003multiple} to solve the scale problem, which has two major shortcomings. First, finding stereo correspondences by individually searching along the respective epipolar lines is computationally expensive. Secondly, if multiple points look similar to the query point, it is challenging to pick the correct one; this happens when the texture is repetitive (\eg grass, sand). We address these two limitations in~\cite{mo2019extending}, where the 3D points in the monocular system are projected into the second camera and the scale problem is solved by minimizing the photometric error. We demonstrate that such direct \textit{scale optimization} is computationally efficient and more robust to repetitive textures in the visual scene.
 
However, even with metric scale, the global camera pose inevitably deviates from the ground truth as the camera moves, because it is estimated by accumulating the relative camera motions incrementally. The \textit{loop closure} brings non-local pose constraints to optimize poses globally to address this issue. The conventional bag-of-word (BoW) approach detects loop closures by matching features from the current view to the history. However, the BoW approach does not work out-of-the-box for direct SLAM systems since direct SLAM systems do not extract descriptors for features. Alternatively, we propose a \emph{LiDAR descriptor-based} place recognition method~\cite{mo2020fast} for urban driving scenarios. We assume the vehicle is moving in the forward direction so that we can accumulate 3D points from the stereo direct SLAM system to imitate LiDAR scans, which are described by a LiDAR descriptor for place recognition. This facilitates significantly more efficient loop closure detection and ensures higher accuracy and robustness. 

In this paper, we systematically combine scale optimization and the LiDAR descriptor-based place recognition approach into a fully-direct stereo SLAM system termed DSV-SLAM; we release an open-source implementation at {{\tt \url{https://github.com/IRVLab/direct_stereo_slam}}}. We conduct thorough experiments to validate its state-of-the-art accuracy, superior computational efficiency, and robustness in visually challenging scenarios. DSV-SLAM demonstrates the feasibility of a full SLAM system without feature detection or matching. In DSV-SLAM, we adopt the state-of-the-art Direct Sparse Odometry (DSO)~\cite{engel2017direct} to track camera poses and estimate 3D points. Then, we extend this to an efficient and accurate stereo visual odometry (VO) using scale optimization~\cite{mo2019extending}. Subsequently, we use the LiDAR descriptor-based place recognition approach~\cite{mo2020fast} to efficiently detect loop closures. The relative poses of potential loop closures are estimated by direct alignment and optionally, further refined by the Iterative Closest Point (ICP) method~\cite{besl1992method}. Finally, we compose and optimize a pose graph to further improve the SLAM accuracy globally. Figure~\ref{fig:kitti00} shows the estimated trajectory and reconstructed environment by DSV-SLAM on sequence $00$ of the KITTI dataset~\cite{geiger2012we}.
\section{Related Work}
\label{sec:related_work}

Visual SLAM has been an active research problem in the robotics and computer vision literature over the last two decades. The early approaches relied on various filter-based estimation methods such as EKF-SLAM~\cite{smith1986representation} and MSCKF\cite{mourikis2007multi}. Starting from PTAM~\cite{klein2007parallel}, many popular approaches incorporate techniques borrowed from structure-from-motion~\cite{hartley2003multiple} (\eg bundle adjustment) into optimization-based visual SLAM systems. The optimization-based visual SLAM systems can be categorized as either feature-based or direct methods depend on whether feature matching is used.

ORB-SLAM~\cite{mur2015orb, mur2017orb, campos2021orb} is one of the most influential and established feature-based methods. In its stereo version~\cite{mur2017orb}, 3D points are triangulated from stereo matching and then tracked across frames. Subsequently, bundle adjustment is applied to jointly optimize the points and camera poses within a local sliding window by minimizing the reprojection error. In the back end, BoW is used for loop closure detection and relative pose estimation. Subsequently, an \textit{essential graph} is optimized to improve the global accuracy. Global bundle adjustment is also executed to further improve the accuracy. Despite the gain in accuracy, it is computationally expensive. 

% LSD-SLAM~\cite{engel2014lsd, engel2015large} is an early development of the direct methods for visual SLAM. When estimating the camera motion, LSD-SLAM minimizes photometric error directly rather than matching features and minimizing the reprojection error. It exhibits higher accuracy and efficiency, especially in poorly-textured (\ie not enough texture or repetitive texture) environments. It is extended to a stereo system in~\cite{engel2015large}, where stereo matching is used for depth initialization. As discussed previously, stereo matching is computationally expensive; it also makes the system no longer a fully direct method. The stereo LSD-SLAM~\cite{engel2015large} calculates photometric error in both cameras, which greatly improves accuracy but also increases computational overhead. Moreover, BoW is no longer applicable for loop closure since the 3D points in LSD-SLAM are not readily selected to be matched across frames. To solve this problem, LSD-SLAM~\cite{engel2014lsd} maintains a specific set of features to be matched across frames using FAB-Map \cite{cummins2008fab}. However, maintaining two individual sets of points (one for direct tracking and one for place recognition) is not very efficient.  

DSO~\cite{engel2017direct, wang2017stereo, gao2018ldso} is the current state-of-the-art for direct visual odometry. Wang~\etal~\cite{wang2017stereo} extend DSO to a stereo system using stereo matching for depth initialization. To incorporate BoW into the DSO system for loop closure, Gao~\etal~\cite{gao2018ldso} modify DSO's point selection strategy to flavor \textit{trackable features} and compute descriptors for these features. However, stereo matching and feature detection and description are computationally expensive and lack robustness to poorly-textured environments.

% Previously, we have proposed scale optimization~\cite{mo2019extending} as an alternative to stereo matching. It solves the unknown scale problem by projecting the 3D points in the monocular VO to the other camera in stereo configuration and estimating the scale by minimizing the photometric error. It achieves comparable accuracy with significantly less computational overhead and offers more robustness to poorly-textured environments. Thus, extending a direct monocular VO to a stereo system using scale optimization retains the advantages of direct methods. Moreover, in~\cite{mo2020fast}, we adapt LiDAR place recognition methods into stereo VO. The proposed system depends on the structure of the environment for place recognition rather than the feature reappearance used in BoW. Therefore, it fits elegantly into a direct stereo VO system. It also runs much faster (see validation in Sec.~\ref{sec:experimental_evaluation}). 

As discussed in Sec.~\ref{sec:introduction}, we proposed scale optimization~\cite{mo2019extending} and LiDAR descriptor-based place recognition~\cite{mo2020fast} as alternatives for stereo matching and BoW approach. They enable a fast and fully direct visual SLAM system, which we attempt to address in this paper.
\section{Methodology}
\label{sec:methodology}

\begin{figure}[ht]
\centering
    \includegraphics[width=\textwidth]{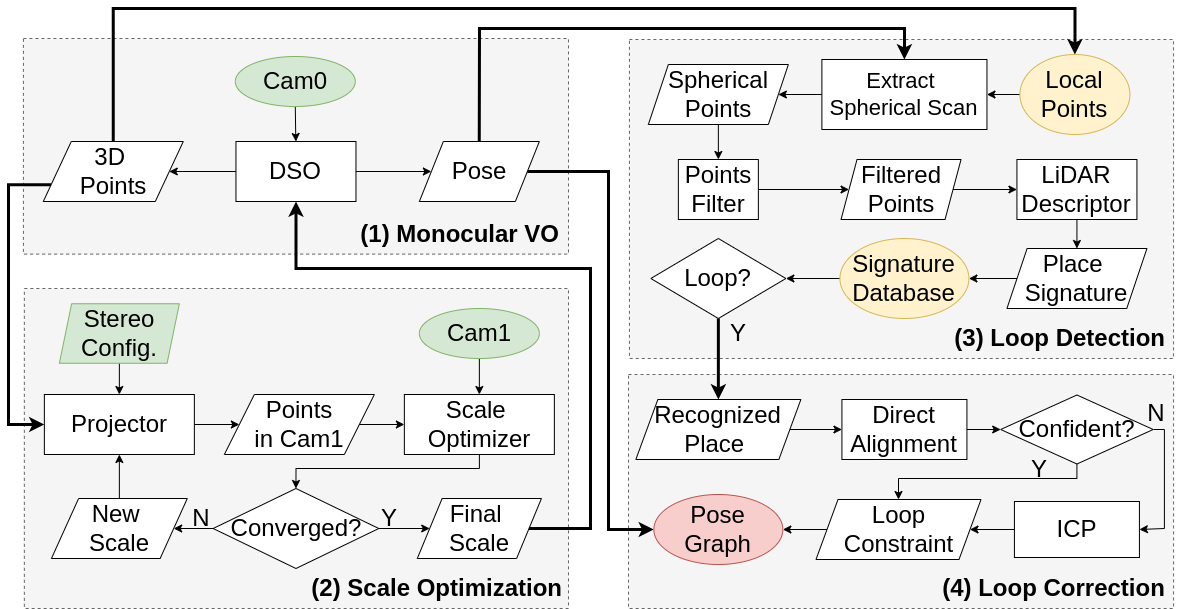}
    \caption{An overview of DSV-SLAM: (1) starting from \textit{Cam0}, the \textit{Monocular VO} estimates camera poses and generates 3D points; (2) using the 3D points, \textit{Scale Optimization} module estimates and maintains the scale of the VO; (3) \textit{Loop Detection} module detects loop closures based on 3D points from VO; (4) for potential loop closures, \textit{Loop Correction} module estimates the relative poses of loop closures and optimizes the poses globally.}
    \label{fig:overview}
\end{figure}

Fig.~\ref{fig:overview} illustrates an outline of the proposed system. There are four computational components: the monocular VO, the scale optimization module, the loop detection module, and the loop correction module.

\subsection*{Notations}
We use ${}_a^b\mathbf{T} = [{}_a^b\mathbf{R}, {}_a^b\mathbf{t}] \in SE(3)$ to represent the \textbf{T}ransformation (\textbf{R}otation and \textbf{t}ranslation) from coordinate $a$ to coordinate $b$. We mark the stereo camera pair as $Cam_0$ and $Cam_1$. For $Cam_k$ where $k \in \{0,1\}$, the corresponding image is $I_k$ and the camera projection is denoted as $\Pi_k$. A 3D point is represented by $\Pi_0^{-1}(\mathbf{p}, d_{\mathbf{p}})$, where $\mathbf{p}$ and $d_{\mathbf{p}}$ are the pixel coordinates and (inverse) depth, respectively, which are back-projected into 3D space by $\Pi_0^{-1}$.

\subsection{Monocular VO}
As mentioned, we choose a direct method over feature-based methods for its accuracy, computational efficiency, and robustness in poorly-textured environments. The current state-of-the-art direct VO method is DSO~\cite{engel2017direct}, which works by minimizing the photometric error defined over a sliding window $\mathcal{F}$ of keyframes and points as

\begin{align}
    \label{eq:dso_error_1}
    E = \sum_{i\in \mathcal{F}}\sum_{\mathbf{p}\in \mathcal{P}_i}\sum_{j\in obs(\mathbf{p})} E_{\mathbf{p}j} \\
    \label{eq:dso_error_2}
    E_{\mathbf{p}j} = \sum_{\mathbf{p}\in \mathcal{N}_\mathbf{p}} \mathit{w}_\mathbf{p} ||(I_0[\mathbf{p}']-b_j) - \frac{t_j e^{a_j}}{t_i e^{a_i}}(I_0[\mathbf{p}]-b_i)||_{\gamma}, \\
    \label{eq:dso_error_3}
    \mathbf{p}' = \Pi_0({}_i^j\mathbf{T}\Pi_0^{-1}(\mathbf{p}, d_{\mathbf{p}})). 
\end{align}

That is, for each point $\mathbf{p} \in \mathcal{P}_i$ in keyframe $i\in \mathcal{F}$, if it is observed by keyframe $j$, then $E_{\mathbf{p}j}$ denotes the associated photometric error. $E_{\mathbf{p}j}$ defined in Eq.~\ref{eq:dso_error_2} is essentially the pixel intensity difference between a point $\mathbf{p}$ in keyframe $i$ and its projection $\mathbf{p'}$ in keyframe $j$ as defined in Eq.~\ref{eq:dso_error_3}; the affine brightness terms ($a_{i/j}$, $b_{i/j}$), exposure times $t_{i/j}$, pixel pattern $\mathcal{N}_p$, weight $\mathit{w}_p$, and Huber norm $||\cdot||_{\gamma}$ are included for photometric robustness. Please refer to~\cite{engel2017direct} for more details. It is worth mentioning that any monocular VO (preferably direct VO) method can be used here instead of DSO due to our modular system design. 

\subsection{Scale Optimization}
As DSO is monocular VO, the scale is unobservable and tends to drift as time goes on. Stereo VO systems solve this problem by bringing the \textit{metric distance} between cameras into the odometry system. As discussed, stereo matching is the conventional way to extend monocular VO to stereo VO but it is computationally expensive and it does not fit well into direct VO. Hence, we adopt scale optimization~\cite{mo2019extending} in the proposed system to balance robustness and efficiency. 

The main idea of scale optimization is to project the points from monocular VO on $Cam_0$ to $Cam_1$ and find the optimal scale that minimizes the photometric error defined as:

\begin{align}
    \label{eq:scale_error_1}
    E = \sum_{\mathbf{p}\in \mathcal{P}}w_{\mathbf{p}} ||I_1[\mathbf{p}']-I_0[\mathbf{p}]||_\gamma, \\
    \label{eq:scale_error_2}
    \mathbf{p}' = \Pi_1({}_0^1\mathbf{R}\cdot s\Pi_0^{-1}(\mathbf{p},d_\mathbf{p}) + {}_0^1\mathbf{t}).
\end{align}

\begin{figure}[ht]
\centering
    \includegraphics[width=\textwidth]{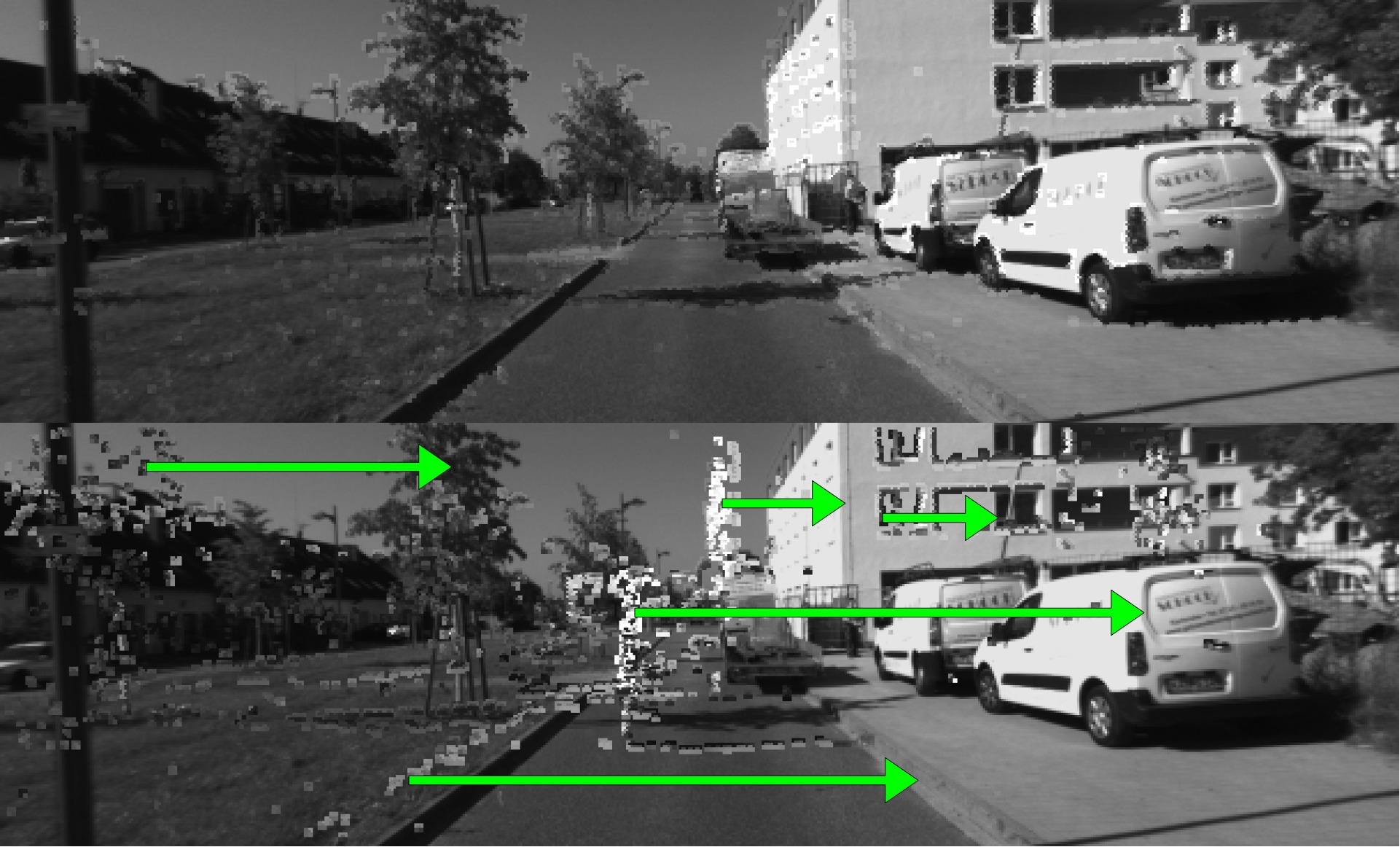}
    \caption{An example of scale optimization on sequence 06 of the KITTI dataset. The top image is the projection with the optimal scale, where the projection well overlaps with the image; the bottom image is the projection with an incorrect scale ($0.1\times$optimal scale), the green arrows indicate the locations of the correct projections.}
    \label{fig:scale_opt}
\end{figure}

For each 3D point $\Pi_0^{-1}(\mathbf{p},d_\mathbf{p})$,  it is re-scaled in $Cam_0$ frame by current scale $s$ and then projected into $Cam_1$ by $[{}_0^1\mathbf{R}, {}_0^1\mathbf{t}]$ and $\Pi_1$, which are known from the stereo calibration. The photometric error $E$ in Eq.~\ref{eq:scale_error_1} is then defined as the pixel intensity difference between the original point $\mathbf{p}$ in $I_0$ and its projection $\mathbf{p}'$ in $I_1$. An example of such scale optimization is illustrated in Fig.~\ref{fig:scale_opt}. Eq.~\ref{eq:scale_error_1} is an analogous formulation of Eq.~\ref{eq:dso_error_2} with two simplifications. First, there is no affine brightness parameters or exposure times. It is feasible as validated in the experiments of \cite{mo2019extending} because a stereo camera is usually hardware synchronized and triggered. Secondly, the photometric error is calculated using a single pixel instead of all pixels in a pattern $\mathcal{N}_p$ (as in Eq.~\ref{eq:dso_error_2}) since the points remain fixed here. Consequently, the scale $s$ is the only \textit{free} parameter to optimize. These simplifications facilitate a computationally efficient optimization process.

Since we do not have prior knowledge of the scale when the system starts, we run scale optimization with a series of initial guesses ranging from $0.1$ to $50$ (empirically chosen) to initialize the scale. Following scale optimization, DSO is adjusted correspondingly by re-scaling the \emph{ Pose} and \emph{ 3D points}. For the consistency of DSO, we only re-scale the pose of the most recently created keyframe and reset its evaluation point; we do not re-scale the other keyframes because of the First Estimate Jacobians~\cite{huang2009first, leutenegger2015keyframe}, but their scale will be optimized heuristically. As a result, the metric scale of DSO is estimated and maintained by scale optimization alone. The resulting stereo VO is computationally efficient and remains fully direct requiring no feature extraction or matching.

\subsection{Loop Detection}
For VO, camera pose drift is inevitable because it is estimated by accumulating local camera motions. To compensate for this error, loop closure brings non-local pose constraints for global pose optimization. BoW~\cite{sivic2003video, galvez2012bags} is the conventional loop closure approach, but it does not fit well into direct methods for reasons discussed previously.

We propose an alternative approach in~\cite{mo2020fast} that fits naturally into direct SLAM. Instead of 2D features, we focus on the 3D structure for place recognition. We adapt LiDAR descriptors on the 3D points from stereo VO to describe a place. However, the 3D points from VO are distributed in a \textit{frustum} due to the narrow field-of-view of cameras. The pose of the frustum changes with that of the camera, which is not desired for place recognition. Our solution to this is illustrated in Fig.~\ref{fig:overview}(3); with the assumption that camera motion is predominantly in the forward direction, we propose to locally accumulate 3D points from VO to get a set of \emph{Local Points}, then generate a set of \emph{Spherical Points} around the current \emph{Pose} to imitate a LiDAR scan. This is feasible because VO is locally accurate. For efficiency, we use \emph{Points Filter} to remove redundant points. The \emph{Filtered Points} make up the final imitated LiDAR scan (\eg Fig.~\ref{fig:icp}). 
To describe the imitated LiDAR scan, we prefer global LiDAR descriptors over local ones mainly for two reasons. First, generating and matching global LiDAR descriptors is usually faster than local ones. Secondly, the imitated LiDAR scan is neither as consistent nor dense as a real LiDAR scan, which is not ideal for local LiDAR descriptors. We are able to use global LiDAR descriptors because the 3D points generated by the proposed stereo VO (DSO with scale optimization) have a metric scale. In \cite{mo2020fast}, we validated that Scan Context~\cite{kim2018scan} is accurate and efficient for datasets recorded in urban areas. Hence, we use Scan Context as our LiDAR descriptor and focus on urban driving scenarios.

\begin{figure}[ht]
\centering
    \includegraphics[width=\textwidth]{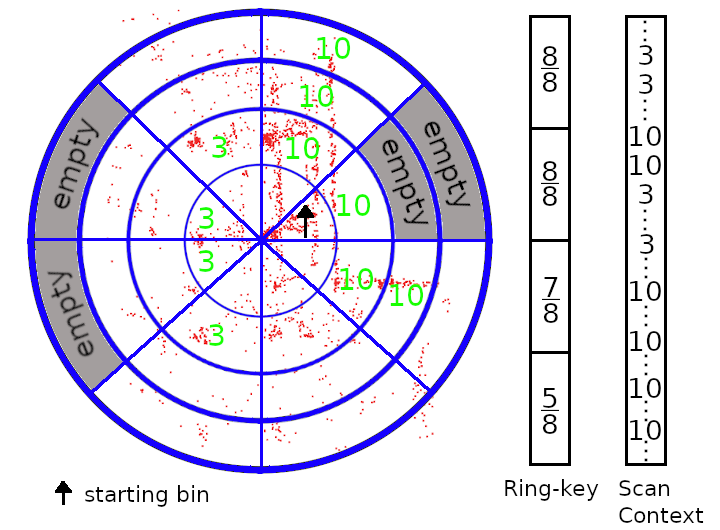}
    \caption{A simplified illustration of ring-key and Scan Context descriptor on the imitated LiDAR scan near the place in Fig.~\ref{fig:scale_opt}. We assume the heights of buildings and trees are $10$ meters and $3$ meters, respectively (for this illustration only).}
    \label{fig:sc}
\end{figure}

The main idea of Scan Context is to use height distribution in urban areas (\eg buildings) to describe the point cloud generated by a LiDAR. The original Scan Context aligns the point cloud with respect to the gravity axis measured by IMU. Since we do not wish to bring additional sensors (\ie IMU) to our visual SLAM system, we align the point cloud using PCA~\cite{tombari2010unique} instead. After alignment, the horizontal plane (the most significant PCA plane in our case) is divided into multiple bins based on \textit{radius} and \textit{azimuth}. The maximal height in each bin is concatenated to form a signature for the current place. The authors of Scan Context also propose to use ring-key~\cite{kim2018scan} for fast preliminary search before Scan Context, which encodes the occupancy ratio in each ring determined by radius. An illustration is given in Fig.~\ref{fig:sc}. 

In our system, for each keyframe from the stereo VO, we imitate a LiDAR scan by the proposed method and generate its place signature using our modified Scan Context descriptor. Then we search for potential loop closure in the \emph{Signature Database}. We first search by ring-key, which is fast but less discriminative, so we select the top three place candidates for Scan Context to make the final decision. 

\subsection{Relative Pose Estimation}
As Fig.~\ref{fig:overview}(4) shows, for each \emph{Recognized Place}, we try to estimate a \emph{Loop Constraint} (\ie the relative pose) between the current place and recognized place. This is achieved by \textit{direct alignment} as done in DSO tracking based on the following equations:

\begin{align}
    \label{eq:pose_error_1}
    E = \sum_{\mathbf{p}\in \mathcal{P}}w_{\mathbf{p}} ||(I_0^c[p']-b_c)-\frac{t_c e^{a_c}}{t_r e^{a_r}}(I_0^r[\mathbf{p}]-b_r)||_\gamma, \\
    \label{eq:pose_error_2}
    p' = \Pi_0({}_r^c\mathbf{T} \Pi_0^{-1}(\mathbf{p},d_\mathbf{p})). 
\end{align}

Here, $I_0^c$ and $I_0^r$ are the \textbf{c}urrent and \textbf{r}ecognized frames, respectively. We are estimating ${}_r^c\mathbf{T}$, the relative pose from recognized frame to current frame, which is initialized by the PCA alignment in \emph{Loop Detection}. The other variables are same as the ones in Eqs.~\ref{eq:dso_error_2} and \ref{eq:dso_error_3}. We specifically project points from the recognized frame to the current frame for memory efficiency, because we only need to store the sparse points instead of the entire image for the recognized frame.

\begin{figure}[ht]
    \centering
    \includegraphics[width=\textwidth]{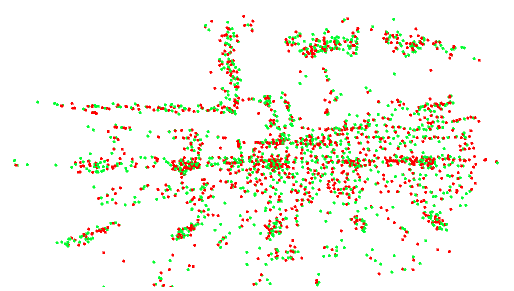}
    \caption{When direct alignment fails, ICP finds the optimal pose that aligns the imitated LiDAR scans of the recognized place (red) and the current place (green).}
    \label{fig:icp}
\end{figure}

Although Eqs.~\ref{eq:pose_error_1} and \ref{eq:pose_error_2} look similar to the error terms in DSO (\ie Eqs.~\ref{eq:dso_error_1}-\ref{eq:dso_error_3}), there are only two keyframes (\ie recognized frame and current frame) in this optimization rather than a sliding window in DSO, and hence, fewer points and constraints; additionally, the illumination, occlusion, and even the scene can vary drastically for loop closures. Consequently, direct alignment alone is not robust enough for loop closure. For robustness, we execute ICP~\cite{besl1992method} to align the imitated LiDAR scans when direct alignment is not confident (Eqs.~\ref{eq:pose_error_1}-\ref{eq:pose_error_2} converge to a large photometric error). An example of ICP is shown in Fig.~\ref{fig:icp}. ICP is particularly robust when the visual appearance changes drastically. Although it is computationally more expensive than direct alignment, the initial relative pose from PCA in \emph{Loop Detection} is reasonably accurate and facilitates a fast convergence. Alternatively, the pose can be estimated by direct alignment and ICP jointly~\cite{park2019robust} for improved accuracy and robustness.

Finally, the \emph{Pose Graph} composed of the consecutive keyframes and loop closures is optimized to improve the pose accuracy globally. Although not implemented yet, global bundle adjustment can be done using the 3D points association from either direct alignment or ICP algorithm for improved map consistency.
\section{Experimental Evaluation}
\label{sec:experimental_evaluation}
To evaluate the accuracy and computational efficiency of DSV-SLAM system, we include several variants of DSO for internal comparison. In particular, we compare the scale optimization in DSV-SLAM with the stereo matching approach adopted in the Stereo DSO\footnote{Since no official release of Stereo DSO is available, we use a third-party implementation in \emph{\url{https://github.com/JingeTu/StereoDSO}}}~\cite{wang2017stereo}. We also compare the performance of our LiDAR descriptor-based place recognition module to the conventional BoW approach used in LDSO~\cite{gao2018ldso}. Externally, we include the performance evaluations of stereo ORB-SLAM2~\cite{mur2017orb} for accuracy and efficiency comparison. 
Since the Scan Context used in this system is designed for the urban driving scenario, we mainly focus on two publicly available datasets: the KITTI visual odometry dataset~\cite{geiger2012we} and the Malaga dataset~\cite{blancoclaraco2014the}. Our experiments are conducted on an Intel\texttrademark{} i7-8750H platform having a 2.2GHz CPU with six cores and 16GB RAM. We use the \textit{default} setting of DSO with $2000$ points located in $5$-$7$ keyframes in the sliding window for optimization (\ie in Eq.~\ref{eq:dso_error_1}-\ref{eq:dso_error_3}). Moreover, when imitating a LiDAR scan for loop detection, we set the LiDAR range (\ie the radius of \emph{Spherical Points} in Fig.~\ref{fig:overview}(3)) to $40$ meters. In the current implementation, scale optimization runs sequentially in the main DSO thread while the loop closure parts (detection, estimation, and pose optimization) run in a separate thread. Due to the inherent randomness in DSO and ORB-SLAM2, we run each algorithm five times and compute the average when calculating accuracy and efficiency.

\subsection{Evaluation on the KITTI Dataset}
The KITTI dataset contains $22$ sequences of stereo images. The ground truths for the first $11$ sequences are publicly available; while the ground truths for the rest are reserved for ranking VO algorithms. We focus on the first $11$ sequences for complete evaluations.

\subsubsection{Accuracy}
To compute accuracy, we align the estimated trajectory to the ground truth and compute the root mean square error of the trajectory as the absolute trajectory error (ATE). Since the DSO and LDSO are monocular systems and unaware of scale, the alignment is based on $Sim3$; Stereo DSO, (stereo) ORB-SLAM2, and DSV-SLAM are aligned with $SE3$. Since the pose graph consists of keyframes only, the comparisons are based on keyframes.

\begin{table}[ht]
    \centering
    \caption{Comparisons for accuracy based on absolute trajectory error (ATE) in \textit{meters} on the KITTI dataset. Results with loop closures are marked with asterisk (*). For Stereo DSO, the results are `official results (3rd implementation)'; for DSV-SLAM, the results are `loop enabled (no loop)'.}
    \begin{tabular}{|l||p{1.2cm}|p{1.2cm}||l|l|p{1.2cm}|}
        %\hline
        \Xhline{2\arrayrulewidth}
        Seq. & DSO ($Sim3$) & LDSO ($Sim3$) & Stereo DSO & DSV-SLAM & ORB-SLAM2 \\
        %\hline
        \Xhline{2\arrayrulewidth}
        $00$ & $102.21$ & $6.51$* & $3.73$ $(4.19)$ & $2.59$* ($5.27$) & $\mathbf{1.19}$*  \\
        \hline
        $01$ & $122.67$ & $7.32$ & $\mathbf{4.07}$ $(5.64)$ & $5.73$ & $9.59$ \\
        \hline
        $02$ & $101.48$ & $15.08$* & $6.99$ $(4.56)$ & $\mathbf{4.18}$* ($5.61$) & $5.35$* \\
        \hline
        $03$ & $1.94$ & $2.07$ & $1.22$ $(1.16)$ & $2.47$ & $\mathbf{0.64}$ \\
        \hline
        $04$ & $0.82$ & $0.95$ & $0.83$ $(0.82)$ & $1.09$ & $\mathbf{0.19}$ \\
        \hline
        $05$ & $46.63$ & $4.48$* & $1.99$ $(2.36)$ & $3.83$* ($3.92$) & $\mathbf{0.72}$* \\
        \hline
        $06$ & $54.21$ & $11.64$* & $1.78$ $(23.57)$ & $0.80$* ($1.04$) & $\mathbf{0.75}$* \\
        \hline
        $07$ & $14.60$ & $13.60$* & $1.15$ $(2.97)$ & $4.37$ & $\mathbf{0.48}$* \\
        \hline
        $08$ & $95.83$ & $100.02$ & $\mathbf{2.70}$ $(3.26)$ & $4.77$ & $3.23$ \\
        \hline
        $09$ & $58.31$ & $60.37$ & $3.63$ $(2.95)$ & $5.32$ & $\mathbf{2.85}$ \\
        \hline
        $10$ & $10.94$ & $13.71$ & $\mathbf{0.77}$ $(1.04)$ & $1.78$ & $1.05$ \\
        %\hline
        \Xhline{2\arrayrulewidth}
    \end{tabular}
    \label{tab:kitti_acc}
\end{table}

\begin{figure}[ht]
    \centering
    \includegraphics[width=\textwidth]{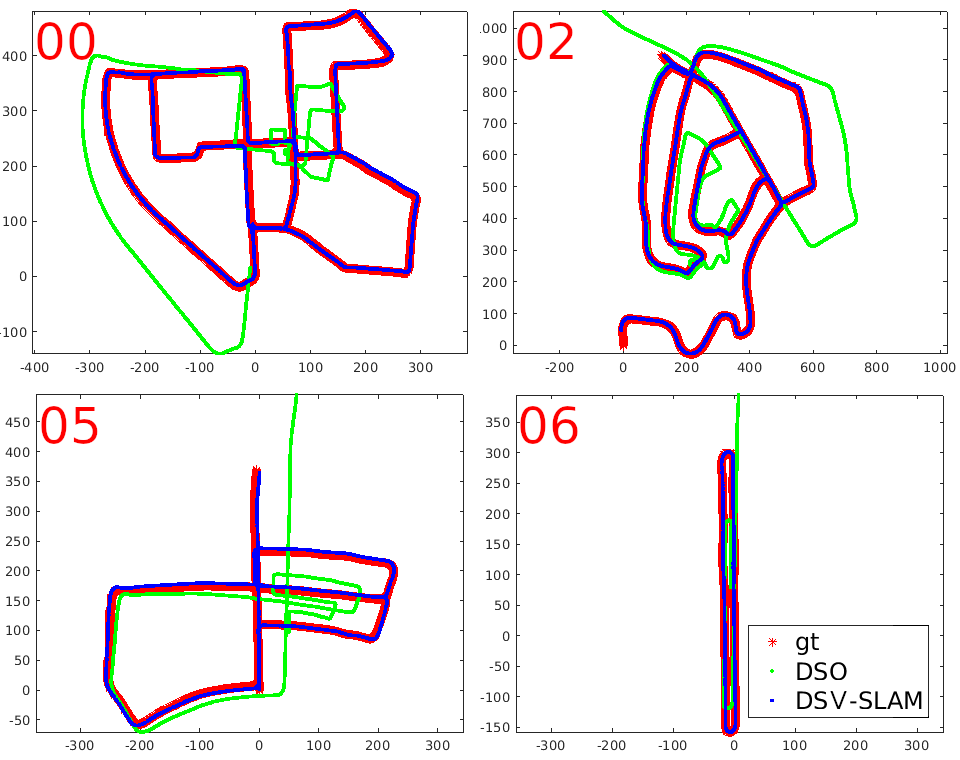}
    \caption{The trajectories estimated by DSO (green), DSV-SLAM (blue), and the ground truth (red) on KITTI sequence $00$, $02$, $05$, and $06$. With scale optimization and loop closure, the improvement of DSV-SLAM over DSO is significant.}
    \label{fig:kitti}
\end{figure}

Table \ref{tab:kitti_acc} reports the accuracy of the state-of-the-art visual SLAM systems on the KITTI dataset. Our results for LDSO and ORB-SLAM2 agree with the results reported in~\cite{gao2018ldso} and ~\cite{mur2017orb}, respectively. The ATEs of Stereo DSO are calculated with the trajectories provided by~\cite{wang2017stereo} (they do not provide code); we also report the results using the 3rd party implementation in parenthesis. 

With DSO being monocular VO, its ATEs are large due to the drifting scale, especially on long sequences like $00$, $02$, and $08$. For LDSO, the ATEs decrease drastically compared to DSO on sequences with loop closures (\ie $00$, $02$, $05$, $06$, and $07$). The scale drifting problem is solved by all the stereo systems. Overall, ORB-SLAM2 performs the best on KITTI dataset, possibly due to the maturity of feature-based methods and the comprehensive system design (\eg global bundle adjustment). For Stereo DSO and DSV-SLAM, although the results on some sequences (\eg $04$) are not as good as ORB-SLAM2, they achieve competitive accuracy on more than half of the sequences. The fast vehicle movement with low camera frame-rate (10Hz) in KITTI dataset are not ideal for direct methods (\ie DSO).

\begin{table}[ht]
\centering
\caption{Comparisons for run-time (\textit{mean} $\times$ \textit{execution count}) on the KITTI dataset. [SM: stereo matching; SO: scale optimization; SC: Scan Context; RK: ring-key; D: direct alignment; I: ICP]}
\label{tab:kitti_time}
\begin{subtable}{\textwidth}
\centering
\caption{Sequence 06 (short)}
\begin{tabular}{|l||l|l|l|l|l|}
    \Xhline{2\arrayrulewidth}
    Methods & DSO & LDSO & \makecell[tl]{Stereo\\DSO} & DSV-SLAM & \makecell[tl]{ORB-\\SLAM2} \\
    \Xhline{2\arrayrulewidth}
    Point Sel. & $4.45$ & $7.16$ & $\mathbf{4.35}$ & $4.43$ & $22.8$ \\
    \hline
    SM/SO & - & - & $10.5$ & $\mathbf{2.09}$ & $17.6$ \\
    \hline 
    BoW/SC & - & $1.71\times828$ & - & $\mathbf{0.46}\times625$ & $7.44\times506$ \\
    \hline 
    Loop Det. & - & $0.33\times828$ & - & \makecell[tl]{$\mathbf{0.05}\times625^{RK}$\\$\mathbf{0.01}\times370^{SC}$} & $6.95\times504$ \\
    \hline 
    Loop Est. & - & $\mathbf{0.44}\times453$ & - & \makecell[tl]{$0.68\times50^D$\\$8.25\times14^I$} & $0.77\times258$ \\
    \hline 
    Pose Opt. & - & $200\times28$ & - & $\mathbf{35.9}\times43.4$ & $589\times1$ \\
    \hline 
    Full BA & - & - & - & - & $9600\times1$ \\
    \hline 
    Loop \# & - & $37.0$ & - & $35.4^D+8.0^I$ & $1.0$ \\
    %\hline 
    \Xhline{2\arrayrulewidth}
\end{tabular}
\end{subtable}

\hfill

\begin{subtable}{\textwidth}
\centering
\caption{Sequence 00 (comprehensive)}
\begin{tabular}{|l||l|l|l|l|l|}
    \Xhline{2\arrayrulewidth}
    Methods & DSO & LDSO & \makecell[tl]{Stereo\\DSO} & DSV-SLAM & \makecell[tl]{ORB-\\SLAM2} \\
    \Xhline{2\arrayrulewidth}
    Point Sel. & $\mathbf{4.72}$ & $8.07$ & $\mathbf{4.72}$ & $4.77$ & $21.2$\\
    \hline
    SM/SO & - & - & $10.8$ & $\mathbf{2.19}$ & $18.7$\\
    \hline 
    BoW/SC & - & $1.89\times3461$ & - & $\mathbf{0.49}\times2321$ & $7.37\times1380$\\
    \hline 
    Loop Det. & - & $1.74\times3461$ & - & \makecell[tl]{$\mathbf{0.06}\times2321^{RK}$\\$\mathbf{0.01}\times1915^{SC}$} & $8.70\times1378$\\
    \hline 
    Loop Est. & - & $\mathbf{0.81}\times2058$ & - & \makecell[tl]{$0.96\times175^D$\\$8.10\times100^I$} & $1.19\times367$\\
    \hline 
    Pose Opt. & - & $1796\times34$ & - & $\mathbf{125}\times110$ & $917\times4$\\
    \hline 
    Full BA & - & - & - & - & $4793\times4$\\
    \hline 
    Loop \# & - & $277.6$ & - & $75^D+35^I$ & $4.0$\\
    %\hline 
    \Xhline{2\arrayrulewidth}
\end{tabular}
\end{subtable}
\end{table}

As the results suggest, the accuracy of DSV-SLAM is comparable to the state-of-the-art visual SLAM systems. With loop closure, the accuracy of DSV-SLAM is further improved on sequences $00$, $02$, $05$, and $06$. Fig.~\ref{fig:kitti} shows the trajectories estimated by DSV-SLAM. Since our stereo VO with scale optimization is already very accurate, the improvement with loop closure is not as drastic as LDSO over DSO. However, unlike LDSO and ORB-SLAM2, DSV-SLAM does not capture the loop closure in sequence $07$. This is because the overlapped trajectory is too short to accumulate \emph{ Local Points} and imitate LiDAR scans for place recognition, whereas BoW works on a single frame.

\subsubsection{Efficiency}
\label{sec:exp_eff}

We investigate the efficiency of each computational component and report the results of a short sequence (06) and a comprehensive sequence (00) in Table~\ref{tab:kitti_time}.

To enable BoW, the \textit{point selection} in LDSO is tuned to prefer features for cross-frame matching, and then a descriptor is extracted for each feature. Consequently, the time spent on point selection is increased compared to DSO. However, the point selection in Stereo DSO and DSV-SLAM is as fast as in DSO. We find that scale optimization (SO) in DSV-SLAM is much faster than the stereo matching (SM) in Stereo DSO and ORB-SLAM2. The stereo matching in ORB-SLAM2 is based on feature descriptors and it is the slowest. On the contrary, in Stereo DSO, the points are projected to the stereo frame and the correspondences are searched around that projection, which is potentially the reason for its faster performance. Nevertheless, scale optimization offers the fastest run-time. 

For loop closure, generating BoW in LDSO is slower than generating the Scan Context (SC) descriptor in DSV-SLAM. Detecting loop closure using BoW is also slower than using the hierarchical search method (\ie ring-key and Scan Context descriptor) in DSV-SLAM. For loop pose estimation, direct alignment in DSV-SLAM is marginally slower than the PnP method~\cite{hartley2003multiple} used in LDSO. Although the ICP in DSV-SLAM is much slower, it is executed only when the direct alignment is not confident, which happens less frequently for simple tests (06). Moreover, the ratio of accepted / proposed loop closures in DSV-SLAM ($\frac{43.4}{50}$ and $\frac{110}{175}$) is much higher than in LDSO ($\frac{37}{453}$ and $\frac{277.6}{2058}$). This indicates that our LiDAR descriptor-based place recognition approach in DSV-SLAM achieves higher precision over the BoW approach (refer to~\cite{mo2020fast} for more detailed validation). Consequently, the time saved by DSV-SLAM on point selection and loop detection is more significant than the loss on loop pose estimation. 

Besides, LDSO spends more time on loop pose optimization; other than consecutive keyframes and loop closures, LDSO also brings the connection between each keyframe and the very first keyframe to the pose graph for accuracy and robustness. Lastly, the loop closure module of ORB-SLAM2 is much slower overall due to its complex mechanisms to improve accuracy and robustness. For instance, ORB-SLAM2 searches the lowest score in its covisibility graph and compares it to the candidate score for loop detection; a loop candidate is accepted only when three consistent and consecutive loop candidates are found in the \textit{covisibility} graph. Such conservative approaches incur considerable computational overhead.

% Compared to the relatively easy sequence $06$, the comprehensive sequence $00$ has a more complex pose graph; hence, pose optimization takes more time to finish. The time on loop detection and loop pose estimation increases more intensively for BoW based approaches (\ie LDSO and ORB-SLAM2) than for DSV-SLAM. However, since the proposed place recognition approach needs to accumulate 3D points locally to imitate a LiDAR scan, DSV-SLAM fails to detect loops if the nearby trajectory does not overlap well. Hence, there are fewer loop closures detected by DSV-SLAM than LDSO on sequence $00$. Nevertheless, the ratio of accepted loop closures over proposed loop closures of DSV-SLAM ($110/175$) is still larger than that of LDSO ($277.6/2058$).

\subsection{Evaluation on the Malaga Dataset}
\begin{figure}
\centering
\includegraphics[width=\textwidth]{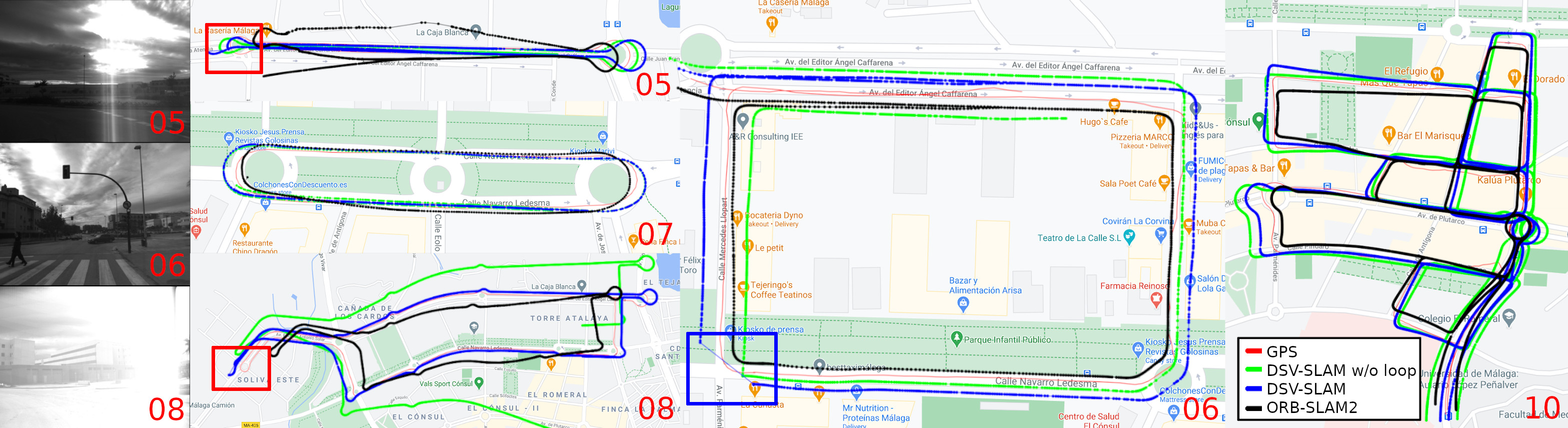}
\caption{Results on Malaga dataset. The blue rectangle in the sequence $06$ shows where the vehicle stops for about $40$ seconds and the underlying DSO in DSV-SLAM loses tracking due to traffic and pedestrians. DSO tracking is also lost in the red rectangles in the sequence $05$ and $08$ due to direct sunlight. Nevertheless, loop closure significantly improves the accuracy of DSV-SLAM in these challenging scenarios. }
\label{fig:malaga}
\end{figure}

To further validate the proposed DSV-SLAM system, we evaluate its performance on the Malaga dataset~\cite{blancoclaraco2014the}. It is more challenging than the KITTI dataset because it consists of various test cases with adverse visual conditions. A few challenging scenarios with poor visibility and direct sunlight are shown in Fig.~\ref{fig:malaga}. In the evaluation, we focus on sequences with loop closures (\ie sequence $05$, $06$, $07$, $08$, and $10$) for testing. Since there is only GPS data available as ground truth, rather than conducting quantitative analyses, we show a qualitative performance comparison in Fig.~\ref{fig:malaga}. Our observations from the experimental results are listed below:
\begin{itemize}
    \item Overall, the scales of the trajectories by both DSV-SLAM and ORB-SLAM2 are slightly inaccurate. We suspect the potential reason is that the structures might be too far for the short-baseline ($12$ cm) stereo camera used in the Malaga dataset.
    
    \item In sequence $05$, the DSO tracking in DSV-SLAM struggles at the roundabout (see the red rectangle in Fig.~\ref{fig:malaga}) due to direct sunlight. Scale optimization also fails several times. However, the shape of DSV-SLAM's trajectory is still more accurate than ORB-SLAM2.
    
    \item In sequence $06$, the DSO tracking in DSV-SLAM also fails due to traffic and pedestrians when the vehicle stops for about $40$ seconds (see the blue rectangle in Fig.~\ref{fig:malaga}). It takes a few seconds to recover tracking, which leads to an inconsistent trajectory estimation by DSV-SLAM without loop closure (denoted by the green trajectory). However, loop closure finds the failure point and eventually corrects the trajectory. ORB-SLAM2 is slightly better with a more accurate scale.
    
    \item In sequence $07$, the orientation of the trajectory estimated by ORB-SLAM2 is slightly off, whereas the scale of DSV-SLAM is slightly off.
    
    \item In sequence $08$, the DSO tracking in DSV-SLAM fails at the red rectangle due to the sudden brightness change. Consequently, the trajectory of DSV-SLAM without loop closure is off; nevertheless, it can re-localize itself with loop closure when the vehicle comes back to the start location. For ORB-SLAM2, the scale of its trajectory is noticeably smaller than the ground truth.
    
    \item Lastly, sequence $10$ is a long run with various straights and turns as well as loop closures, which tests visual SLAM algorithms comprehensively. The trajectory generated by DSV-SLAM is slightly more accurate than ORB-SLAM2. We also notice that the distance between the start and the end of the trajectory is considerably reduced by the loop closure (from the green trajectory to the blue one).
\end{itemize}

Overall, we find that DSV-SLAM's accuracy is comparable and often better than ORB-SLAM2 on the Malaga dataset. However, DSV-SLAM is computationally more efficient with significant margins as presented in Table \ref{tab:malaga_time}. ICP is executed more frequently on the Malaga dataset than on the KITTI dataset since direct alignment is vulnerable to brightness change.

\begin{table}[ht]
\centering
\caption{Comparisons for run-time (\textit{mean millisecond} $\times$ \textit{execution count}) on sequence 10 of the Malaga dataset. [SM: stereo matching; SO: scale optimization; SC: Scan Context; RK: ring-key; D: direct alignment; I: ICP]}
\label{tab:malaga_time}
\centering
\begin{tabular}{|l||l|l|}
%\hline
\Xhline{2\arrayrulewidth}
 & DSV-SLAM & ORB-SLAM2 \\
%\hline
\Xhline{2\arrayrulewidth}
Select Point & $\mathbf{7.50}\times5053$ & $28.7\times17310$ \\
\hline
SM/SO & $\mathbf{4.06}\times5050$ & $24.6\times17310$ \\
\hline
BoW/SC Gen. & $\mathbf{0.64}\times3890$ & $7.76\times3072$ \\
\hline
Loop Det. & $\mathbf{0.07}\times3890^{RK}+\mathbf{0.04}\times3369^{SC}$ & $15.3\times3070$ \\
\hline
Loop Est. & $1.91\times854^D+11.98\times755^I$ & $\mathbf{2.09}\times1129$ \\
\hline
Pose Opt. & $\mathbf{164.2}\times215$ & $4202\times11$ \\
\hline 
Full BA & - & $90578\times11$ \\
\hline
FPS & $\mathbf{27.5}$ & $14.0$ \\
\hline
Loop Count & $99^D+116.4^I$ & $11.6$ \\
\Xhline{2\arrayrulewidth}
\end{tabular}
\end{table}

\subsection{Evaluation on the RobotCar Dataset}
\begin{figure}[ht]
    \centering
    \includegraphics[width=\textwidth]{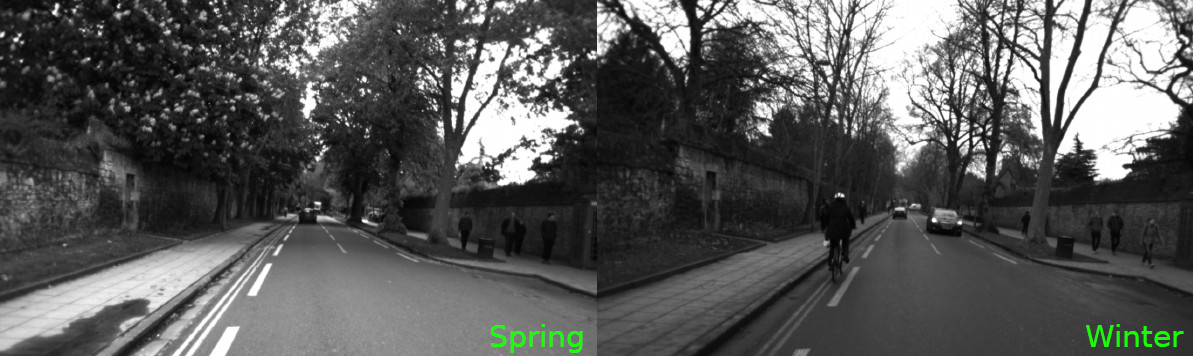}
    \caption{Snapshots of the RobotCar dataset. There are many visual appearance differences including trees and foliage, traffic, pedestrians, and varying brightness.}
    \label{fig:robotcar}
\end{figure}

\begin{figure}[ht]
    \centering
    \includegraphics[width=\textwidth]{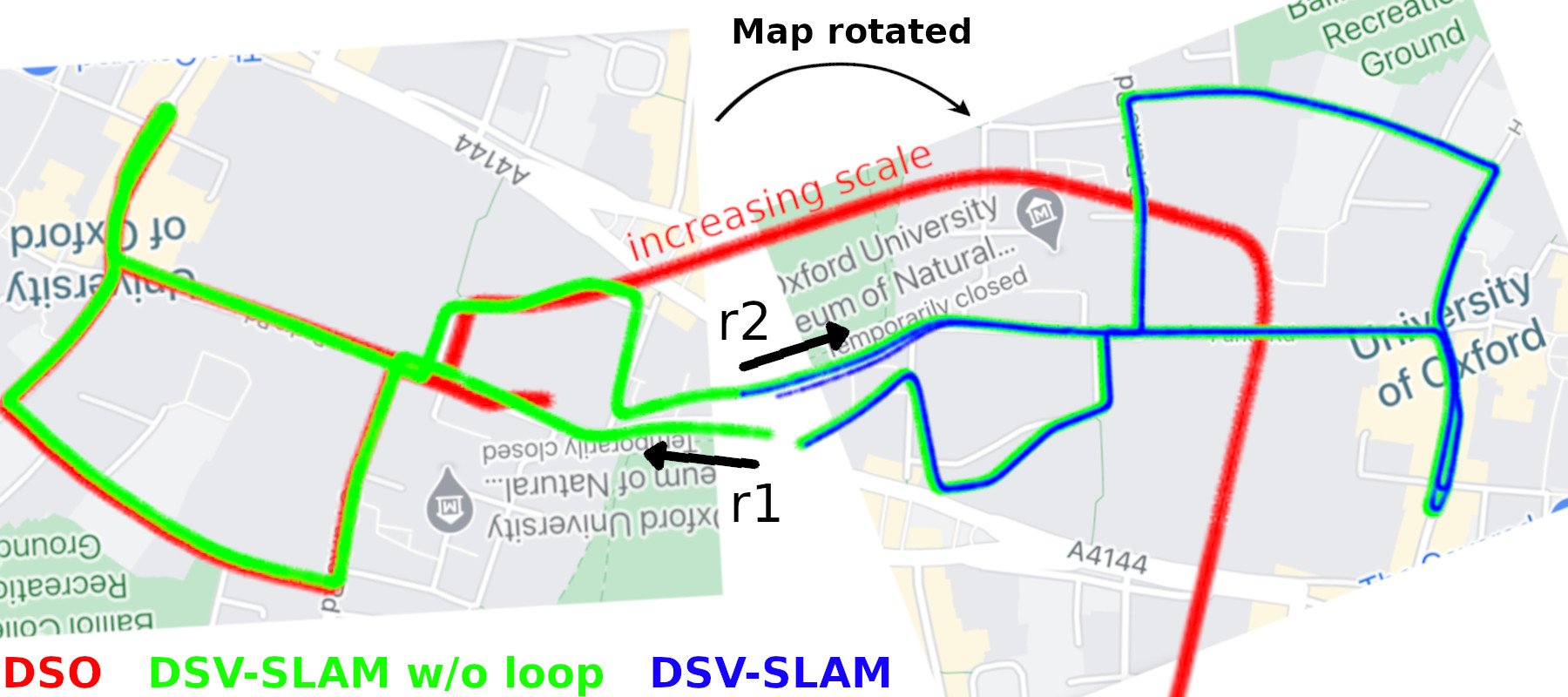}
    \caption{The trajectories estimated by DSO, DSV-SLAM with and without loop closure on RobotCar dataset. While DSO has an increasing scale issue, DSV-SLAM estimates the scale accurately and consistently, and it is able to re-localize the robot after being kidnapped using loop closures. [r1: start of run1; r2: start of run2 where the robot is kidnapped to]}
    \label{fig:robotcar_traj}
\end{figure}

RobotCar dataset~\cite{maddern20171} is recorded in different seasons throughout the year, which we used to validate the robustness of the LiDAR descriptor-based place recognition approach against visual appearance changes in \cite{mo2020fast}. Snapshots are given in Fig.~\ref{fig:robotcar}. We demonstrate the preliminary result of DSV-SLAM on the RobotCar dataset in Fig. \ref{fig:robotcar_traj}, where we first play the sequence `2015-05-19-14-06-38' (\textit{run1}), and then we ``kidnap'' the robot to sequence `2015-08-13-16-02-58'(\textit{run2}). As Fig. \ref{fig:robotcar_traj} shows, the DSO scale gets larger throughout; the drifting scale is fixed with scale optimization (see the green trajectory); with loop closures, the robot eventually re-localize itself and bring the two runs together (see the blue trajectory). We also run ORB-SLAM2 using the same setting; however, its tracking fails consistently. 

\section{Conclusions}
In this paper, we propose the first fully-direct visual SLAM system for driving scenarios, demonstrating the feasibility of a full SLAM system without feature detection or matching. We first extend the monocular DSO to a stereo system using scale optimization; then we integrate a LiDAR descriptor-based place recognition approach to detect loop closures; for potential loop closures, we use direct alignment to estimate the relative pose, which is bolstered by ICP when the direct alignment is not confident. Validation on public datasets demonstrates that the proposed system achieves considerably better computational efficiency while offering comparable accuracy and improved robustness in challenging scenarios. For future work, we will look into eliminating the forward-moving camera assumption when imitating LiDAR scans to expand our potential use cases. We also intend to extend the system into a stereo-visual-inertial system by integrating IMU measurements to further improve robustness.

\section{Acknowledgement}
This work was supported by US National Science Foundation Award IIS \#1637875, University of Minnesota Doctoral Dissertation Fellowship, and MnRI Seed Grant.
%We gratefully acknowledge the support of the MnDRIVE initiative for this research. 

\bibliographystyle{plain}
\bibliography{main}

\begin{thebibliography}{10}

\bibitem{besl1992method}
Paul~J Besl and Neil~D McKay.
\newblock {A Method for Registration of 3-D Shapes}.
\newblock In {\em Sensor Fusion IV: Control Paradigms and Data Structures},
  volume 1611, pages 586--606. International Society for Optics and Photonics,
  1992.

\bibitem{blancoclaraco2014the}
Joseluis Blancoclaraco, Franciscoangel Morenoduenas, and Javier
  Gonzalezjimenez.
\newblock {The Málaga Urban Dataset: High-rate Stereo and Lidars in a
  Realistic Urban Scenario}.
\newblock {\em The International Journal of Robotics Research}, 33(2):207--214,
  2014.

\bibitem{bresson2017simultaneous}
Guillaume Bresson, Zayed Alsayed, Li~Yu, and S{\'e}bastien Glaser.
\newblock {Simultaneous Localization and Mapping: A Survey of Current Trends in
  Autonomous Driving}.
\newblock {\em IEEE Transactions on Intelligent Vehicles}, 2(3):194--220, 2017.

\bibitem{cadena2016past}
Cesar Cadena, Luca Carlone, Henry Carrillo, Yasir Latif, Davide Scaramuzza,
  Jos{\'e} Neira, Ian Reid, and John~J Leonard.
\newblock {Past, Present, and Future of Simultaneous Localization And Mapping:
  Towards the Robust-Perception Age. Simultaneous Localization and Mapping}.
\newblock {\em IEEE Transactions on Robotics}, 32(6):1309--1332, 2016.

\bibitem{campos2021orb}
Carlos Campos, Richard Elvira, Juan J.~Gómez Rodríguez, José~M. M.~Montiel,
  and Juan D.~Tardós.
\newblock {ORB-SLAM3: An Accurate Open-Source Library for Visual,
  Visual–Inertial, and Multimap SLAM}.
\newblock {\em IEEE Transactions on Robotics}, pages 1--17, 2021.

\bibitem{engel2017direct}
Jakob Engel, Vladlen Koltun, and Daniel Cremers.
\newblock {Direct Sparse Odometry}.
\newblock {\em IEEE Transactions on Pattern Analysis and Machine Intelligence},
  40(3):611--625, 2017.

\bibitem{engel2014lsd}
Jakob Engel, Thomas Sch{\"o}ps, and Daniel Cremers.
\newblock {LSD-SLAM: Large-Scale Direct Monocular SLAM}.
\newblock In {\em European Conference on Computer Vision}, pages 834--849.
  Springer, 2014.

\bibitem{engel2015large}
Jakob Engel, J{\"o}rg St{\"u}ckler, and Daniel Cremers.
\newblock {Large-Scale Direct SLAM with Stereo Cameras}.
\newblock In {\em 2015 IEEE/RSJ International Conference on Intelligent Robots
  and Systems (IROS)}, pages 1935--1942. IEEE, 2015.

\bibitem{forster2016svo}
Christian Forster, Zichao Zhang, Michael Gassner, Manuel Werlberger, and Davide
  Scaramuzza.
\newblock {SVO: Semidirect Visual Odometry for Monocular and Multicamera
  Systems}.
\newblock {\em IEEE Transactions on Robotics}, 33(2):249--265, 2016.

\bibitem{fuentes2015visual}
Jorge Fuentes-Pacheco, Jos{\'e} Ruiz-Ascencio, and Juan~Manuel
  Rend{\'o}n-Mancha.
\newblock {Visual Simultaneous Localization and Mapping: A Survey}.
\newblock {\em Artificial Intelligence Review}, 43(1):55--81, 2015.

\bibitem{galvez2012bags}
Dorian G{\'a}lvez-L{\'o}pez and Juan~D Tardos.
\newblock {Bags of Binary Words for Fast Place Recognition in Image Sequences}.
\newblock {\em IEEE Transactions on Robotics}, 28(5):1188--1197, 2012.

\bibitem{gao2018ldso}
Xiang Gao, Rui Wang, Nikolaus Demmel, and Daniel Cremers.
\newblock {LDSO: Direct Sparse Odometry with Loop Closure}.
\newblock In {\em 2018 IEEE/RSJ International Conference on Intelligent Robots
  and Systems (IROS)}, pages 2198--2204. IEEE, 2018.

\bibitem{geiger2012we}
Andreas Geiger, Philip Lenz, and Raquel Urtasun.
\newblock {Are We Ready for Autonomous Driving? The KITTI Vision Benchmark
  Suite}.
\newblock In {\em 2012 IEEE Conference on Computer Vision and Pattern
  Recognition}, pages 3354--3361. IEEE, 2012.

\bibitem{grasa2013visual}
Oscar~G Grasa, Ernesto Bernal, Santiago Casado, Ismael Gil, and JMM Montiel.
\newblock {Visual SLAM for Handheld Monocular Endoscope}.
\newblock {\em IEEE transactions on medical imaging}, 33(1):135--146, 2013.

\bibitem{hartley2003multiple}
Richard Hartley and Andrew Zisserman.
\newblock {\em {Multiple View Geometry in Computer Vision}}.
\newblock Cambridge University Press, 2003.

\bibitem{huang2009first}
Guoquan~P Huang, Anastasios~I Mourikis, and Stergios~I Roumeliotis.
\newblock {A First-Estimates Jacobian EKF for Improving SLAM Consistency}.
\newblock In {\em Experimental Robotics}, pages 373--382. Springer, 2009.

\bibitem{kim2018scan}
Giseop Kim and Ayoung Kim.
\newblock {Scan Context: Egocentric Spatial Descriptor for Place Recognition
  within 3D Point Cloud Map}.
\newblock In {\em 2018 IEEE/RSJ International Conference on Intelligent Robots
  and Systems (IROS)}, pages 4802--4809. IEEE, 2018.

\bibitem{klein2007parallel}
Georg Klein and David Murray.
\newblock {Parallel Tracking and Mapping for Small AR Workspaces}.
\newblock In {\em Proceedings of the 2007 6th IEEE and ACM International
  Symposium on Mixed and Augmented Reality}, pages 1--10. IEEE Computer
  Society, 2007.

\bibitem{leutenegger2015keyframe}
Stefan Leutenegger, Simon Lynen, Michael Bosse, Roland Siegwart, and Paul
  Furgale.
\newblock {Keyframe-based Visual-Inertial Odometry using Nonlinear
  Optimization}.
\newblock {\em The International Journal of Robotics Research}, 34(3):314--334,
  2015.

\bibitem{maddern20171}
Will Maddern, Geoffrey Pascoe, Chris Linegar, and Paul Newman.
\newblock {1 Year, 1000km: The Oxford RobotCar Dataset}.
\newblock {\em The International Journal of Robotics Research}, 36(1):3--15,
  2017.

\bibitem{mo2019extending}
Jiawei Mo and Junaed Sattar.
\newblock {Extending Monocular Visual Odometry to Stereo Camera Systems by
  Scale Optimization}.
\newblock In {\em 2019 IEEE/RSJ International Conference on Intelligent Robots
  and Systems (IROS)}, pages 6921--6927, 2019.

\bibitem{mo2020fast}
Jiawei Mo and Junaed Sattar.
\newblock {A Fast and Robust Place Recognition Approach for Stereo Visual
  Odometry Using LiDAR Descriptors}.
\newblock In {\em 2020 IEEE/RSJ International Conference on Intelligent Robots
  and Systems (IROS)}, pages 5893--5900, 2020.

\bibitem{mourikis2007multi}
Anastasios~I Mourikis and Stergios~I Roumeliotis.
\newblock {A Multi-State Constraint Kalman Filter for Vision-aided Inertial
  Navigation}.
\newblock In {\em Proceedings 2007 IEEE International Conference on Robotics
  and Automation}, pages 3565--3572. IEEE, 2007.

\bibitem{mur2015orb}
Raul Mur-Artal, Jose Maria~Martinez Montiel, and Juan~D Tardos.
\newblock {ORB-SLAM: A Versatile and Accurate Monocular SLAM System}.
\newblock {\em IEEE Transactions on Robotics}, 31(5):1147--1163, 2015.

\bibitem{mur2017orb}
Raul Mur-Artal and Juan~D Tard{\'o}s.
\newblock {ORB-SLAM2: An Open-Source SLAM System for Monocular, Stereo, and
  RGB-D Cameras}.
\newblock {\em IEEE Transactions on Robotics}, 33(5):1255--1262, 2017.

\bibitem{park2019robust}
Chanoh Park, Soohwan Kim, Peyman Moghadam, Jiadong Guo, Sridha Sridharan, and
  Clinton Fookes.
\newblock {Robust Photogeometric Localization Over Time for Map-Centric Loop
  Closure}.
\newblock {\em IEEE Robotics and Automation Letters}, 4(2):1768--1775, 2019.

\bibitem{sivic2003video}
Josef Sivic and Andrew Zisserman.
\newblock {Video Google: A Text Retrieval Approach to Object Matching in
  Videos}.
\newblock In {\em Proceedings of the IEEE International Conference on Computer
  Vision}, pages 1470--1478, 2003.

\bibitem{smith1986representation}
Randall~C Smith and Peter Cheeseman.
\newblock {On the Representation and Estimation of Spatial Uncertainty}.
\newblock {\em The International Journal of Robotics Research}, 5(4):56--68,
  1986.

\bibitem{thrun2007simultaneous}
Sebastian Thrun.
\newblock {Simultaneous Localization and Mapping}.
\newblock In {\em Robotics and cognitive approaches to spatial mapping}, pages
  13--41. Springer, 2007.

\bibitem{tombari2010unique}
Federico Tombari, Samuele Salti, and Luigi Di~Stefano.
\newblock {Unique Signatures of Histograms for Local Surface Description}.
\newblock In {\em European Conference on Computer Vision}, pages 356--369.
  Springer, 2010.

\bibitem{wang2017stereo}
Rui Wang, Martin Schworer, and Daniel Cremers.
\newblock {Stereo DSO: Large-Scale Direct Sparse Visual Odometry with Stereo
  Cameras}.
\newblock In {\em Proceedings of the IEEE International Conference on Computer
  Vision}, pages 3903--3911, 2017.

\bibitem{weiss2011monocular}
Stephan Weiss, Davide Scaramuzza, and Roland Siegwart.
\newblock {Monocular-SLAM-based Navigation for Autonomous Micro Helicopters in
  GPS-denied Environments}.
\newblock {\em Journal of Field Robotics}, 28(6):854--874, 2011.

\end{thebibliography}
\end{document}